\documentclass[conference]{IEEEtran}
\IEEEoverridecommandlockouts
\usepackage{amsmath,amssymb,amsfonts}
\usepackage{lettrine}
\usepackage{booktabs}
\usepackage{algorithm,algpseudocode}
\usepackage{graphicx}
\usepackage{subfigure}
\usepackage{multirow}
\usepackage{textcomp}
\usepackage{xcolor}
\usepackage{svg}
\usepackage[colorlinks=true,linkcolor=blue, citecolor=blue,urlcolor=black]{hyperref}%
\usepackage{orcidlink}
\usepackage[doi=false, isbn=false, style=ieee]{biblatex} 
\AtBeginBibliography{\footnotesize}
\bibliography{references.bib}

\algnewcommand{\Inputs}[1]{%
  \State \textbf{Input:}
  \Statex \hspace*{\algorithmicindent}\parbox[t]{.8\linewidth}{\raggedright #1}
}
\algnewcommand{\Parameters}[1]{%
  \State \textbf{Parameters:}
  \Statex \hspace*{\algorithmicindent}\parbox[t]{.8\linewidth}{\raggedright #1}
}
\algnewcommand{\Embeddings}[1]{%
  \State \textbf{Initial Embeddings:}
  \Statex \hspace*{\algorithmicindent}\parbox[t]{.8\linewidth}{\raggedright #1}
}
\algnewcommand{\Initialize}[1]{%
  \State \textbf{Initialize:}
  \Statex \hspace*{\algorithmicindent}\parbox[t]{.8\linewidth}{\raggedright #1}
}
\algnewcommand{\GAP}[1]{%
  \State \textbf{Global Average Pooling:}
  \Statex \hspace*{\algorithmicindent}\parbox[t]{.8\linewidth}{\raggedright #1}
}
\algnewcommand{\Classifier}[1]{%
  \State \textbf{Classifier:}
  \Statex \hspace*{\algorithmicindent}\parbox[t]{.8\linewidth}{\raggedright #1}
}
\algnewcommand{\Output}[1]{%
  \State \textbf{Output:}
  \Statex \hspace*{\algorithmicindent}\parbox[t]{.8\linewidth}{\raggedright #1}
}

\begin{document}

\twocolumn[
\begin{@twocolumnfalse}
	\Huge {IEEE copyright notice} \\ \\
	\large {\copyright\ 2024 IEEE. Personal use of this material is permitted. Permission from IEEE must be obtained for all other uses, in any current or future media, including reprinting/republishing this material for advertising or promotional purposes, creating new collective works, for resale or redistribution to servers or lists, or reuse of any copyrighted component of this work in other works.} \\ \\
	
	{\Large \textbf{Accepted to be Published in :} \emph{3rd IEEE International Conference on
   Computer Vision and Machine Intelligence (IEEE CVMI)}, October 19 - 20, 2024, IIIT Allahabad, Prayagraj, India.} \\
	
\end{@twocolumnfalse}
]

\title{StrideNET: Swin Transformer for Terrain Recognition with Dynamic Roughness Extraction}

\author{
Maitreya~Shelare\textsuperscript{\dag\thanks{\textsuperscript{\dag}Corresponding Author}}
\quad Neha~Shigvan
\quad Atharva~Satam
\quad Poonam~Sonar\\
{Rajiv Gandhi Institute of Technology, University of Mumbai, India} \\
\small{\texttt{\{maitreya.cse,nehatshigvan,atharvajsatam17\}@gmail.com, poonam.sonar@mctrgit.ac.in}}
}
\maketitle
\begin{abstract}
The field of remote-sensing image classification has seen immense progress with the rise of convolutional neural networks, and more recently, through vision transformers. These models, with their self-attention mechanism, can effectively capture global relationships and long-range dependencies between the image patches, in contrast with traditional convolutional models. This paper introduces StrideNET, a dual-branch transformer-based model developed for terrain recognition and surface roughness extraction. The terrain recognition branch employs the Swin Transformer to classify varied terrains by leveraging its capability to capture both local and global features. Complementing this, the roughness extraction branch utilizes a  statistical texture-feature analysis technique to dynamically extract important land surface properties such as roughness and slipperiness. The model was trained on a custom dataset consisting of four terrain classes — grassy, marshy, sandy, and rocky, and it outperforms benchmark CNN and transformer based models, by achieving an average test accuracy of over $99 \%$ across all classes. The applications of this work extend to different domains such as environmental monitoring, land use and cover classification, disaster response and precision agriculture.

\end{abstract}


\section{Introduction} \label{introduction}

\lettrine[lines=2]{T}{errain} recognition and extraction of its properties such as roughness \& slipperiness, by fusion of deep learning and remote sensing techniques offers meaningful benefits across diverse domains. These applications include land use and land cover (\textit{LULC}) classification~\cite{8848484}, ecological monitoring~\cite{WILLIS2015233}, geographical mapping~\cite{https://doi.org/10.1111/gcb.13388}, natural feature detection~\cite{doi:10.1080/13658816.2018.1542697}, and disaster management~\cite{HOQUE2017345}. 

Traditionally, terrain recognition was done manually by experts, which was a time-consuming and expensive process. To automate this, different image processing techniques were proposed~\cite{10.5555/3137503} earlier. But they all failed to classify rapidly changing terrain accurately. This drawback was later overcome by using deep learning techniques for classification. 

The Convolutional Neural Networks (CNN) stands as one of the most extensively used deep learning technique. CNN based methods excel at classifying images with high accuracy even in very challenging situations~\cite{9324261}. 

However, their locality bias~\cite{rs15071860} limits them from capturing long-range dependencies and global relationships within the image and they often lack the ability to explain the rationale behind their inferences~\cite{10.1145/2939672.2939778}. 

These limitations of CNN-based methods are overcome by the Vision Transformer~\cite{dosovitskiy2021image} and its  variants~\cite{touvron2021training}, which have shown promising results in numerous computer vision tasks.

Thus, a novel dual-branch transformer-based model is proposed, named 
StrideNET: Swin Transformer for Terrain Recognition with Dynamic Roughness Extraction.

The \textit{Terrain Recognition} branch uses the Swin Transformer to classify different terrains. Swin Sransformer is a variant of the Vision Transformer architecture, which constructs hierarchical representation of the input image. By utilizing shifted window-based self-attention, it also establishes cross-window connections while maintaining computationally efficient local window computations.

By restricting computations to non-overlapping local windows, Swin Transformer achieves a linear time complexity of $O(mn)$, unlike the complexity of $O(n^2)$ offered by the traditional the Vision Transformer, if the window size $m$ is kept reasonably small~\cite{liu2021swin}. Moreover, it offers higher generalization capabilities over CNNs by considering the relationship between different features of an image~\cite{rs14020359}.

The \textit{Roughness Extraction} branch uses a statistical texture-feature analysis technique to dynamically extract surface properties like roughness and slipperiness, by determining how pixels interact within local areas of an image by capturing changes in the grayscale levels.

It utilizes statistical methods to model surface texture as a random field and then fits a probability distribution to the intensity distribution within that texture ~\cite{Bhuyan2020-li}. Using this, first the variance of each image patch is calculated and then the corresponding roughness factor is computed.  

The key contributions of this work are summarized below:

\begin{enumerate}
        \item A novel algorithm for extraction land surface properties such as roughness and slipperiness is proposed, which utilizes statistical texture-feature analysis for inference.
        \item The StrideNET model achieves exceptional classification accuracy in terrain recognition, surpassing other benchmark models and further validating its effectiveness.
        \end{enumerate}

The following is the structure of this paper. An overview of related work in terrain recognition and roughness extraction is provided in Section \ref{related work}. The proposed StrideNET model is explained in Section \ref{method}. Experiments \& Results are discussed in Section \ref{experimental results}. A summary of our work is given in Section \ref{conclusion}.

\begin{figure*}
    \centering
    \includegraphics[width=18 cm]{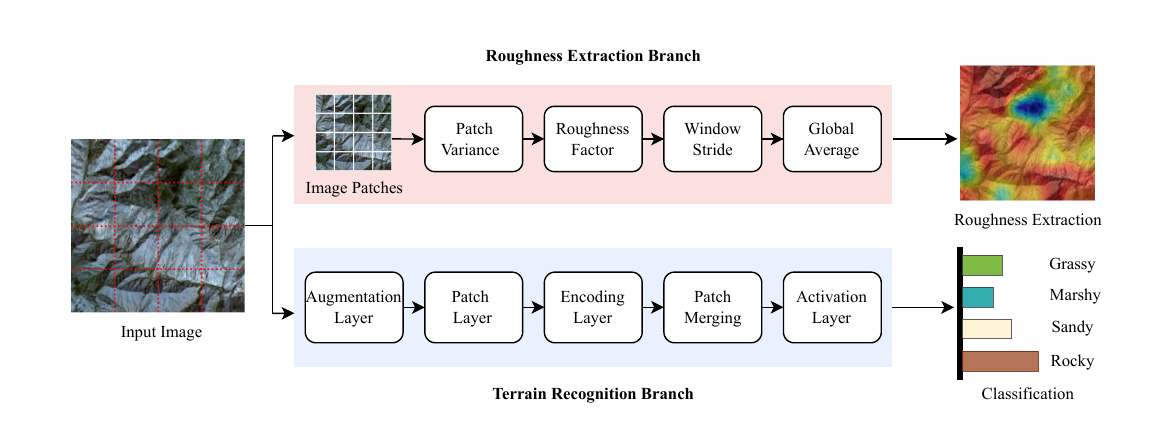}
    \caption{StrideNET Architecture}
    \label{fig:architecture}
\end{figure*}

\section{Related Works}\label{related work}

Over the years, numerous deep learning techniques have been proposed for terrain recognition. The Faster-RCNN model~\cite{10.1145/3149808.3149814}, uses a deep convolutional neural network (D-CNN) for accurate detection of craters in aerial and remote sensing imagery, although it is limited to only binary classification. 

W. Li et al.~\cite{Li2020-jf} conducts more advanced experiments for classification of natural terrain features, by comparing multiple convolutional models, and reporting that Inception-ResNet hybrid model outperforms other traditional CNN models. 

Z. Yu~\cite{9236884} explores the fusion of AlexNet with K-nearest neighbors algorithm in remote sensing terrain classification, remarking that the hybrid method yields improved performance compared to other methods.

A. A. Aleissaee et al.~\cite{Aleissaee2023-vc} reviews the performance of various transformer models across different remote sensing tasks, reporting that these models outperform their convolutional counterparts due to their ability to effectively capture long-range dependencies within images.

The TRS model~\cite{zhang} combines traditional convolutional neural networks with vision transformers, by replacing their spatial convolutions with multi-head self-attention, which leads to significant improvements in classification performance.

Y. Bazi et al.~\cite{bazi} demonstrates that the attention mechanism of vision transformers can be used to capture the contextual relations between different image patches effectively, which results in better classification accuracy. 

V. Suryamurthy et al.~\cite{Suryamurthy} propose a deep neural network using SegNet and ERFNet for pixel-wise terrain labeling and roughness prediction, leveraging low-level CNN features and up-projection blocks to restore spatial resolution.

Z. Yu~\cite{YuZ} presents a self-supervised model using a whiskered robot to capture vibrations for terrain classification and roughness estimation with low computational cost.

\section{Proposed Model - Stridenet}\label{method}

\begin{figure*}
    \centering
    \includegraphics[width=18 cm]{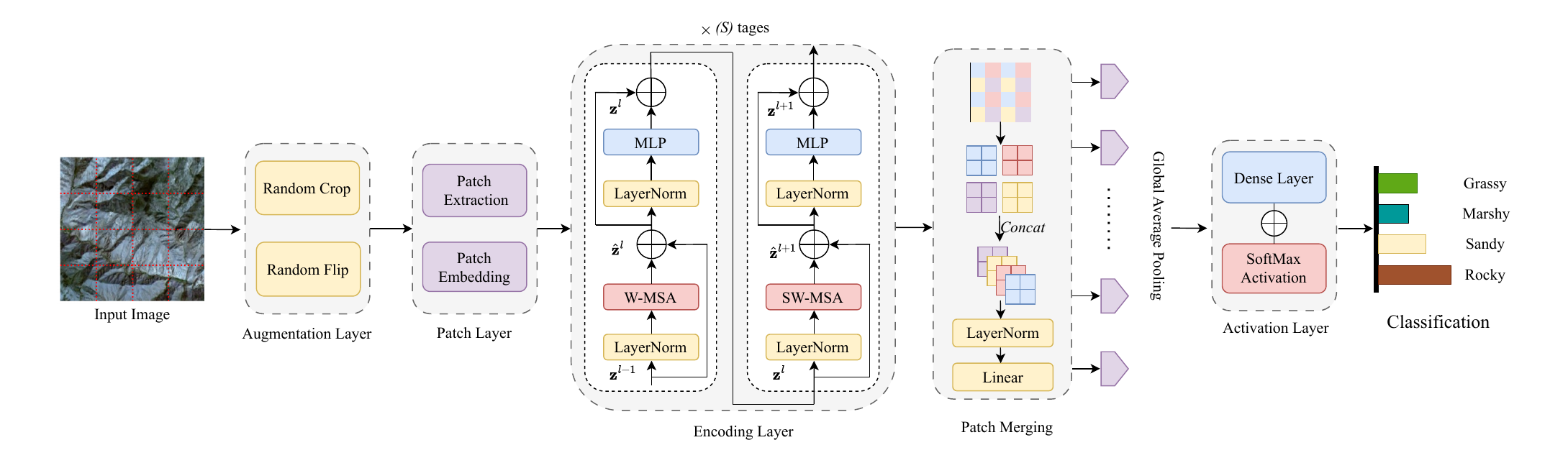}
    \caption{Terrain Recognition branch}
    \label{fig:classification branch}
\end{figure*}

The model architecture is illustrated in Fig. \ref{fig:architecture}, where the input image is processed through two distinct branches: the \textit{Terrain Recognition} branch and the \textit{Roughness Extraction} branch.

\subsection{Mathematical Background}

The StrideNET model is built using the Swin Transformer architecture, which differs from standard Vision Transformer in three key aspects, which are discussed below:

\subsubsection{Self-attention in non-overlapped windows}

Swin Transformer utilizes a hierarchical structure using self-attention for enabling efficient processing of high-resolution images. 

The shifted window self-attention further improves the capacity of the model to capture long-range dependencies by introducing cross-window connections. 

Multi-head self-attention ($\mathcal{MSA}$) is a mechanism designed to capture long-range dependencies among pixels in an image, and it is computed using Eq. \ref{eq1}.

\begin{equation}
\Omega(\mathcal{MSA})=4 h w C^2+2(h w)^2 C\label{eq1}
\end{equation}

The windowed multi-head self-attention ($\mathcal{W-MSA}$) block in the Swin Transformer is a more computationally efficient variant of $\mathcal{MSA}$. The formula for $\mathcal{W-MSA}$, which takes four inputs, is provided in Eq. \ref{eq2}.

\begin{equation}
\Omega(\mathcal{W-MSA})=4 h w C^2+2(M)^2 h w C\label{eq2}
\end{equation}

In Eq. \ref{eq1} and Eq. \ref{eq2}, ${h}$ and ${w}$ represent the height and width of the feature maps, ${C}$ denotes the number of channels in the input feature map, and ${M}$ is the number of attention heads.

\subsubsection{Shifted window partition in successive blocks}

Swin Transformer adopts a shifted windowing approach which confines self-attention to disjoint windows while enabling cross-window connectivity. 

By alternating the partitioning configuration across blocks, this technique enhances the model's ability to capture long-range dependencies and global context within images more effectively.

\begin{equation}
\begin{array}{l}{{\hat{\bf z}^{l}=\mathcal{W-MSA}\left(\mathcal{LN}\left({\bf z}^{l-1}\right)\right)+{\bf z}^{l-1},}} \\ 

{{{\bf z}^{l}=\mathcal{MLP}\left(\mathcal{LN}\left({\hat{\bf z}}^{l}\right)\right)+{\hat{\bf z}}^{l},}} \\ 

{{\hat{\bf z}^{l+1}=\mathcal{SW-MSA}\left(\mathcal{LN}\left({\bf z}^{l}\right)\right)+{\bf z}^{l},}}\\ 

{{{\bf z}^{l+1}=\mathcal{MLP}\left(\mathcal{LN}\left({\hat{\bf z}}^{l+1}\right)\right)+{\hat{\bf z}}^{l+1},}}\end{array} \label{eq3}
\end{equation}

In Eq. \ref{eq3}, the $\mathcal{W-MSA}(\cdot)$ operation applies self-attention within local windows by partitioning the input feature map into disjoint regions, while $\mathcal{LN}(\cdot)$ denotes layer normalization.

The $\mathcal{MLP}(\cdot)$ consists of two fully connected layers, each incorporating a $\mathrm{GELU}$ activation function, positioned between the $\mathcal{SW-MSA}(\cdot)$ operations. The variable ${\bf z}$ represents the feature map at each stage of the Swin Transformer.

\subsubsection{Relative position bias}

 Swin Transformer employs relative positional bias to improve the model's performance. This technique allows the attention mechanism to more effectively focus on different segments of the input sequence by considering their relative positions. 
\begin{equation}
\mathrm{Attention}(Q,K,V)=\mathrm{SoftMax}(Q K^{T}/\sqrt{d}+B)V \label{eq4}
\end{equation}

where ${Q}$, ${K}$ and ${V}$ are the query, key and value vectors respectively. ${B}$ is the relative position bias matrix \& ${d}$ denotes the dimension of the key vector.

The relative position bias matrix ${B}$ is a learned matrix that encodes the relative positions of elements in the input sequence. It enhances attention mechanisms by enabling more effective learning of how to attend to different parts of the sequence.

The StrideNET model leverages the capabilities of the Swin Transformer to achieve accurate terrain recognition.

\subsection{Terrain Recognition Branch}\label{SWIN Transformer}

\begin{algorithm}
  \caption{Swin Transformer for Terrain Recognition}
  \label{algorithm:recognition}
  \begin{algorithmic}[1]
    \Inputs{Image: $\textbf{\textit{I}}\in\mathbb{R}^{224\times224\times3}$}
    
    \Parameters{Patch size: $P = 4 \times 4$ \\Embedded Dimension: $D = 96$\\
    Heads in MSA: $H$\\Transformer Blocks: $T = \{2,2,6,2\}$\vspace{0.1cm}} 
    
    \Embeddings{Split $\textit{I}$ into patches: 
    $ I_p\in\mathbb{R}^{56\times56\times48}$\\
    Embed: $E_1 = Linear(I_p) ; E_1\in\mathbb{R}^{56 \times 56 \times D}$\vspace{0.1cm}}  

    \For{$s = 2$ to $4$}
      \State \textbf{a. Patch Merge:}
        \State $E_s = \text{Merge}(E_{s-1})$
      \State \textbf{b. Process:}
        \For{$t = 1$ to $T[s]$}
          \State $E_s = \text{TransformerBlock}(E_s)$
        \EndFor
    \EndFor

    \GAP{$V = GAP(E_4); V\in\mathbb{R}^D$}
    \Classifier{$O=Softmax(Linear(V))$}
    \Output{$O$: Class Probabilities of Different Terrains}
  \end{algorithmic}
\end{algorithm}

The \textit{Terrain Recognition} branch is illustrated in Fig. \ref{fig:classification branch}. The input image $I$ is first processed by an augmentation layer that applies operations such as cropping and flipping to enhance the model's robustness. Next, a patch extraction process divides the image into patches of size $P$, forming a three-dimensional tensor $I_p$ that represents distinct local regions. 

Positional embedding is then utilized to encode the spatial information of the patches. This is achieved by applying a linear transformation to each image patch, and embedding it into a \textit{D}-dimensional vector, resulting in a new tensor \textit{E}. The image patches, along with their corresponding positional embeddings, are then fed into the encoding layer of the Swin Transformer.

\begin{figure*}
\centering  
\subfigure[Grassy]{\includegraphics[width=0.47\linewidth]{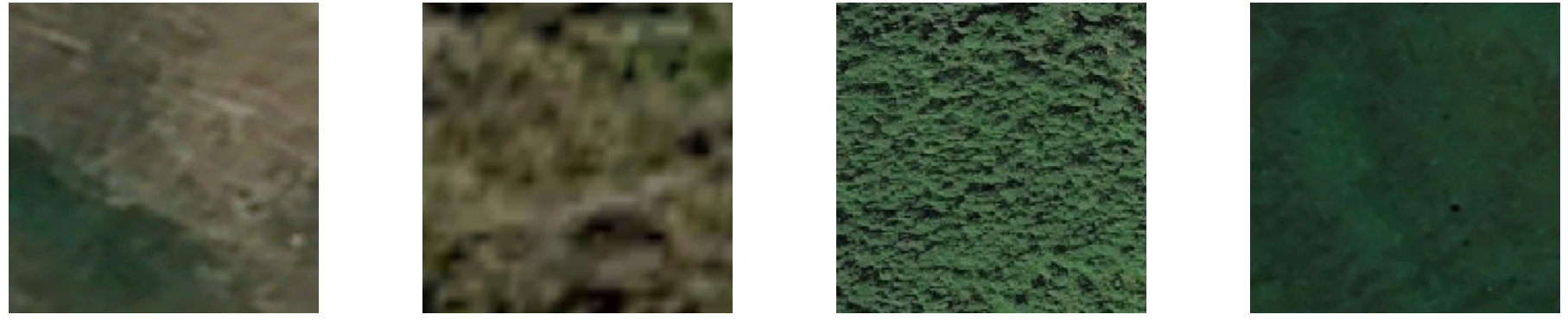}}\hfill
\subfigure[Marshy]{\includegraphics[width=0.47\linewidth]{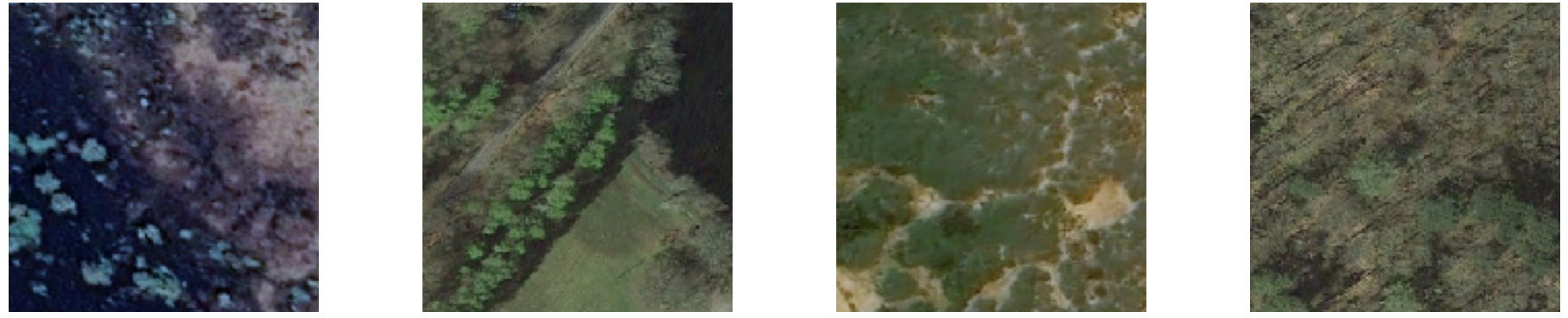}}\hfill
\subfigure[Rocky]{\includegraphics[width=0.47\linewidth]{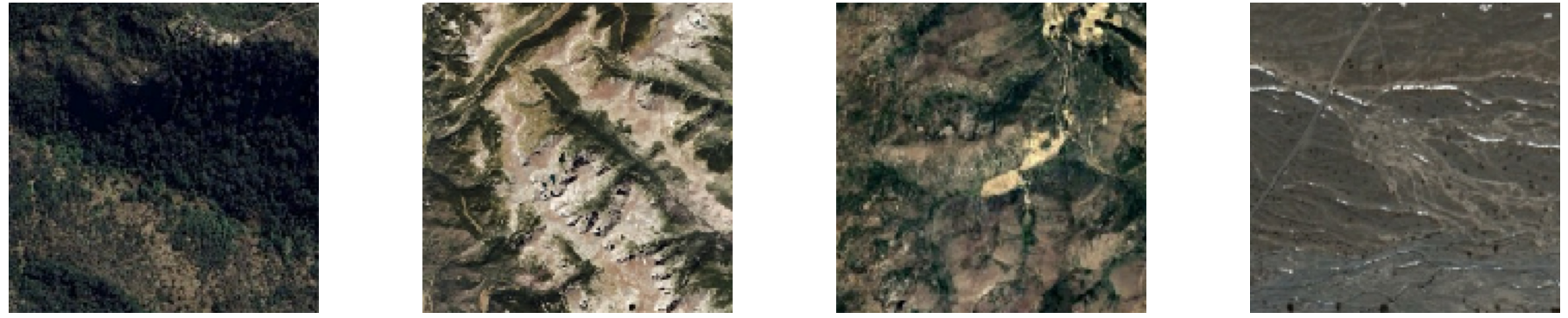}}\hfill
\subfigure[Sandy]{\includegraphics[width=0.47\linewidth]{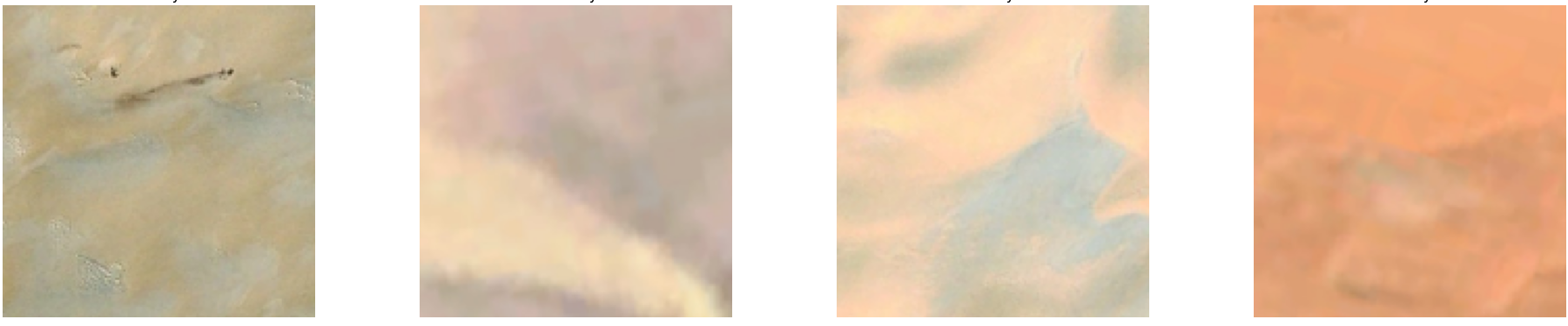}}\hfill
\caption{Terrain Dataset~\cite{Aras_2023}}
\label{fig:dataset}
\end{figure*}

The encoding layers comprise successive transformer blocks, each equipped with multiple attention heads and a feedforward neural network. The shifted windowed self-attention mechanism is used to efficiently capture long-range dependencies within the image. The output is subsequently passed through the feedforward neural network for feature transformation.

Following this, patch merging is performed to downsample the feature maps between each stage of the transformer, except at the final stage. 

Next, global average pooling is applied to compress the spatial information from the feature maps into a fixed-length vector, ensuring a uniform input size.

The output $V$ from the global average pooling layer is forwarded to a dense layer, which maps the high-dimensional feature representation to the desired output dimension of four, corresponding to the number of terrain classes. 

Finally, Softmax activation is used to assign probability to each terrain class. This branch is summarized in Algorithm \ref{algorithm:recognition}.

\begin{figure*}
    \centering
    \includegraphics[width=16 cm]{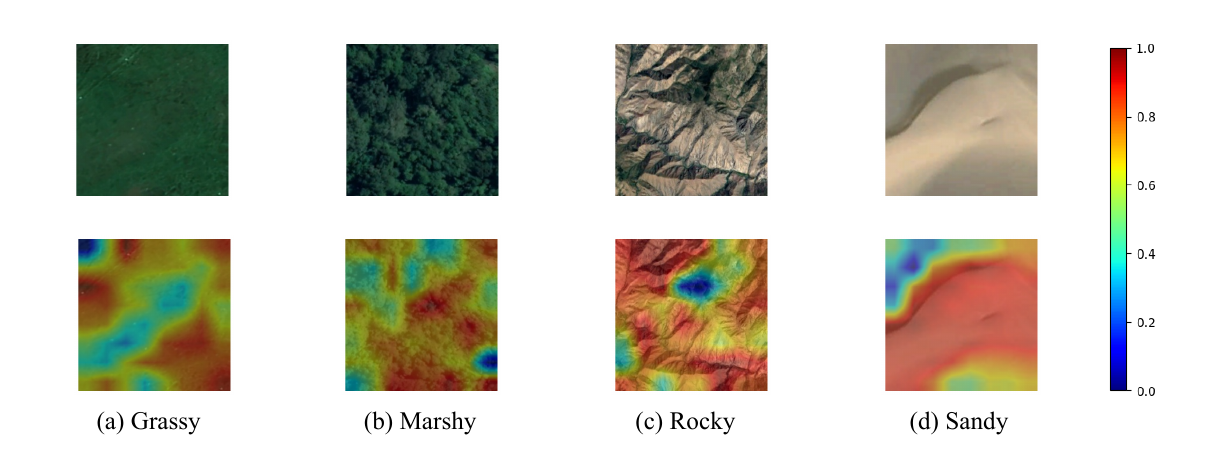}
    \caption{Roughness Extraction}
    \label{fig:roughness estimation}
\end{figure*}

\subsection{Roughness Extraction Branch}\label{Statistical Texture Analysis}

The \textit{Roughness Extraction} branch uses different image characteristics like texture and variance to compute the value of roughness factor $R$.  

Texture describes the recurring pattern of localized intensity variations in an image. It quantifies how these intensities are arranged within a specific region and is usually represented as a feature vector. Statistical methods are particularly effective for analyzing small texture elements, which contribute to microtextures.

Variance measures the spread of intensity values within an image. It is a dimensionless indicator of how much these values deviate from the mean intensity. A high variance indicates a wide distribution of intensity values, characteristic of high-contrast images. Conversely, low variance indicates that intensity values are closely clustered, typical of low-contrast images.

 \begin{algorithm}
\caption{Terrain Roughness Extraction}
\label{algorithm:roughness}
\begin{algorithmic}[1]

\Inputs{Image: $\textbf{\textit{I}}\in\mathbb{R}^{224\times224\times3}$}

\Parameters{Patch size: $P = W \times H$ \\
            Step Size: $s$\\
            \vspace{0.1cm}}

    \Procedure{ImplicitProperties}{$I, s$}
        \State $P \gets \text{Patchify}(I, P, s)$
        \For{$P \in Patches$}
            \State $PatchList \gets PatchList \cup \{P\}$
            \State $Window Stride \gets s$
        \EndFor
        \For{$V \in Variance(PatchList)$}
            \State $Roughness \gets Roughness \cup \{1 - \frac{1}{1 + V}\}$
        \EndFor
            \EndProcedure
            \State $GlobalRoughness \gets \text{GlobalAverage}(Roughness)$

    \Procedure{Visualize}{$GlobalRoughness$}        
        \State $Data \gets \text{CreateMatrix}(GlobalRoughness)$
        \State $Data \gets \text{ColorMap}(Data)$
        \State $O \gets \text{Blend}(I, Data)$
        \State \Return $O$
        \EndProcedure

    \Output{$O$: Terrain image with extracted properties}

\end{algorithmic}
\end{algorithm}

Variance to be computed from the image histogram is given by Eq. \ref{eq5}

\begin{equation}
\sigma^{2}=\sum_{i=0}^{j-1}(z_{i}-m)^{2}\,p(z_{i}) \label{eq5}
\end{equation}

 where  ${z_{i}}$ is the intensity value of the  $i{\textsuperscript{th}}$ pixel,  ${m}$ denotes the mean intensity value \&  $p (z_{i})$ is the probability of the $i^{\text{th}}$ pixel having intensity value ${z_{i}}$.

The roughness factor ${R}$ is a measure of the texture of an image, which is computed using Eq. \ref{eq6}  

\begin{equation}
R=1-\frac{1}{1+\sigma^{2}} \label{eq6}
\end{equation}

It is a dimensionless quantity ranging from 0 to 1, where 0 represents a completely smooth image and 1 represents an entirely rough image ~\cite{Bhuyan2020-li}.

The \textit{Roughness Extraction} branch, outlined in Algorithm \ref{algorithm:roughness}, uses the roughness factor ${R}$ to dynamically extracts roughness and slipperiness from the image.

First, the input image \textit{I} is segmented into patches of size \textit{P}, creating a collection of patches known as \textit{PatchList}. The algorithm then processes each patch in \textit{PatchList} by calculating its variance to determine its roughness value. These roughness values are stored in the \textit{Roughness} matrix. The global roughness value is subsequently obtained by averaging the roughness values from the \textit{Roughness} matrix, and this average is recorded in the \textit{GlobalRoughness} matrix.

To visualize the extracted roughness value, a \textit{Data} matrix is generated. This matrix is normalized to a common range and resized to match the dimensions of the original image. A colormap is then applied to depict varying roughness levels. The final output \textit{O} is obtained by blending the original image \textit{I} with the matrix, as illustrated in Fig. \ref{fig:roughness estimation}.

\section{Experiments and Results}\label{experimental results}

\subsection{Dataset Description and Training Details}

The model is trained on a custom dataset consisting of over $45,000$ images, with approximately $10,000$ images per class for each terrain type: Sandy, Rocky, Grassy, and Marshy, as shown in Fig. \ref{fig:dataset}. This dataset is publicly available at~\cite{Aras_2023}.

All experiments are performed using PyTorch on a system equipped with an Intel Core $i7-11800H$ CPU, $16$ GB DDR$4$ RAM, and an Nvidia GeForce RTX $3050$ Ti GPU.

For model training, the input image is resized from $256 \times 256$ to $224 \times 224$. The patch size is set to $4$, and window size is configured to $7$. Label Smoothing Cross-Entropy loss is used as the objective function, with AdamW as optimizer.

The model undergoes training for a total of $10$ epochs, with a $70:30$ data split. The learning rate is initialized to $0.001$, and a StepLR scheduler is applied with a step size of $3$ epochs and a decay factor of $0.97$.

To prevent the model from overfitting, a dropout rate of $0.3$ is applied. The model architecture consists of transformer layers configured as \([2, 2, 6, 2]\) and attention heads configured as \([3, 6, 12, 24]\).

\begin{table}[t]
\centering
\caption{Classwise Accuracy (in \%) of different methods on the Terrain Recognition dataset.}
\label{tab:performance comparision}
\resizebox{3.4in}{!}{
\begin{tabular}{ccccccc}
\toprule
Class & MobileNet-V2 & EfficientNet-B0 & ResNet-101 & ViT & DeiT & \textbf{StrideNET }  \\ 

\midrule

Grassy      & 95.98   & 95.42 & 97.69 & 97.54 & 99.37 & 99.56 \\
Marshy      & 93.83   & 94.01 & 98.32 & 96.28 & 98.93 & 99.05 \\
Rocky       & 97.89   & 93.87 & 98.44 & 97.00 & 99.29 & 99.20 \\
Sandy       & 94.30   & 94.32 & 97.03 & 98.17 & 98.54 & 99.93 \\ 

\bottomrule
\end{tabular}}
\end{table}

\begin{table}[t]
\centering
\caption{Performance Metrics (in \%) of different methods on the Terrain Recognition dataset.}
\label{tab:metrics}
\resizebox{3.4in}{!}{
\begin{tabular}{ccccccc}
\toprule
Methods          & OA    & AA    & Kappa & Precision & Recall & F1 Score    \\ \midrule
MobileNet-V2    & 95.30 & 95.50 & 90.83 & 75.50     & 65.32  & 70.30 \\
EfficientNet-B0 & 95.02 & 94.40 & 93.42 & 95.02     & 92.30  & 94.30 \\
ResNet-101      & 96.83 & 97.87 & 95.94 & 96.32     & 95.69  & 96.69 \\
ViT             & 98.61 & 97.25 & 98.84 & 98.61     & 99.03  & 99.03 \\
DeiT            & 99.86 & 99.03 & 99.18 & 99.86     & 98.49  & 99.49 \\
\textbf{StrideNET}       & 99.98 & 99.41 & 99.96 & 99.98     & 99.95  & 99.99 \\ \bottomrule
\end{tabular}}
\end{table}

\begin{figure}
\centering  
\subfigure[Accuracy Curve]{\includegraphics[width=0.5\linewidth]{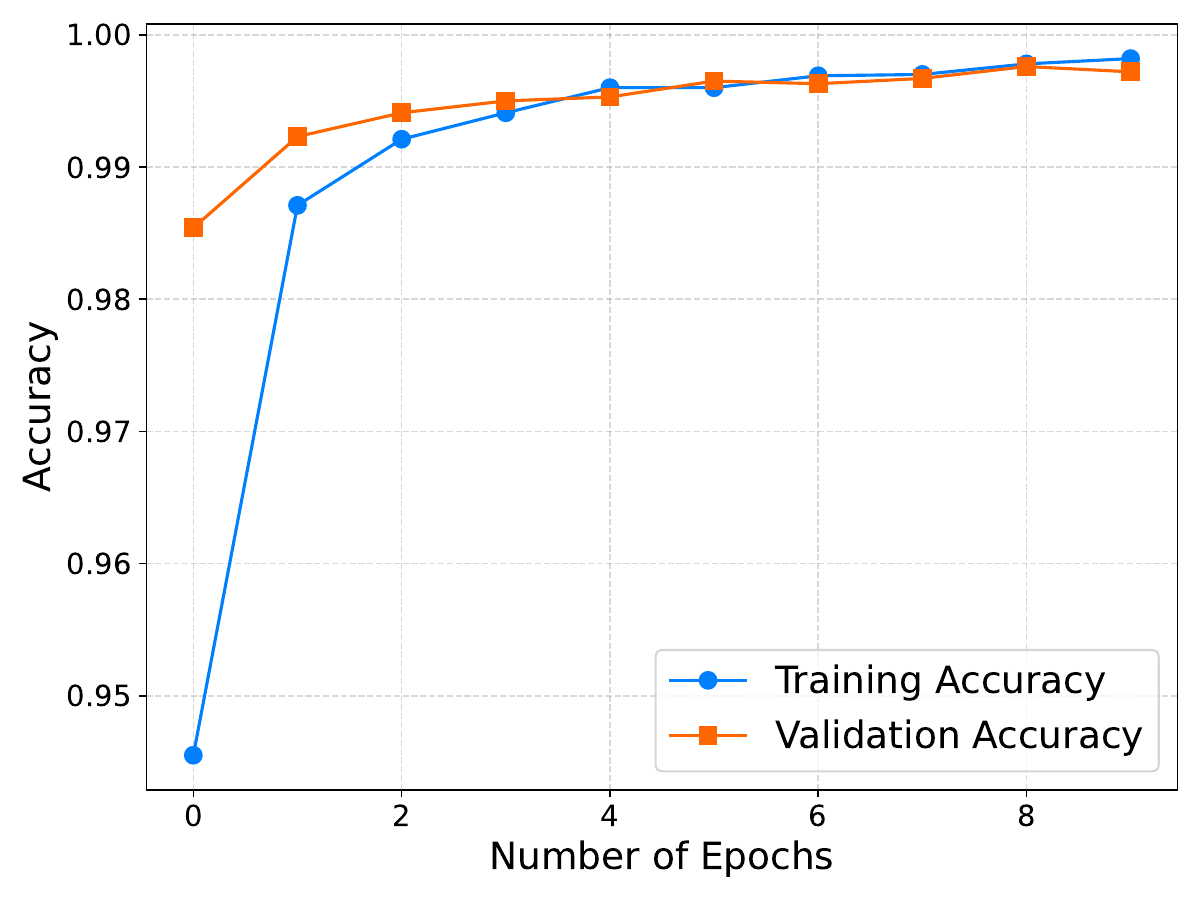}}\hfill
\subfigure[Loss Curve]{\includegraphics[width=0.5\linewidth]{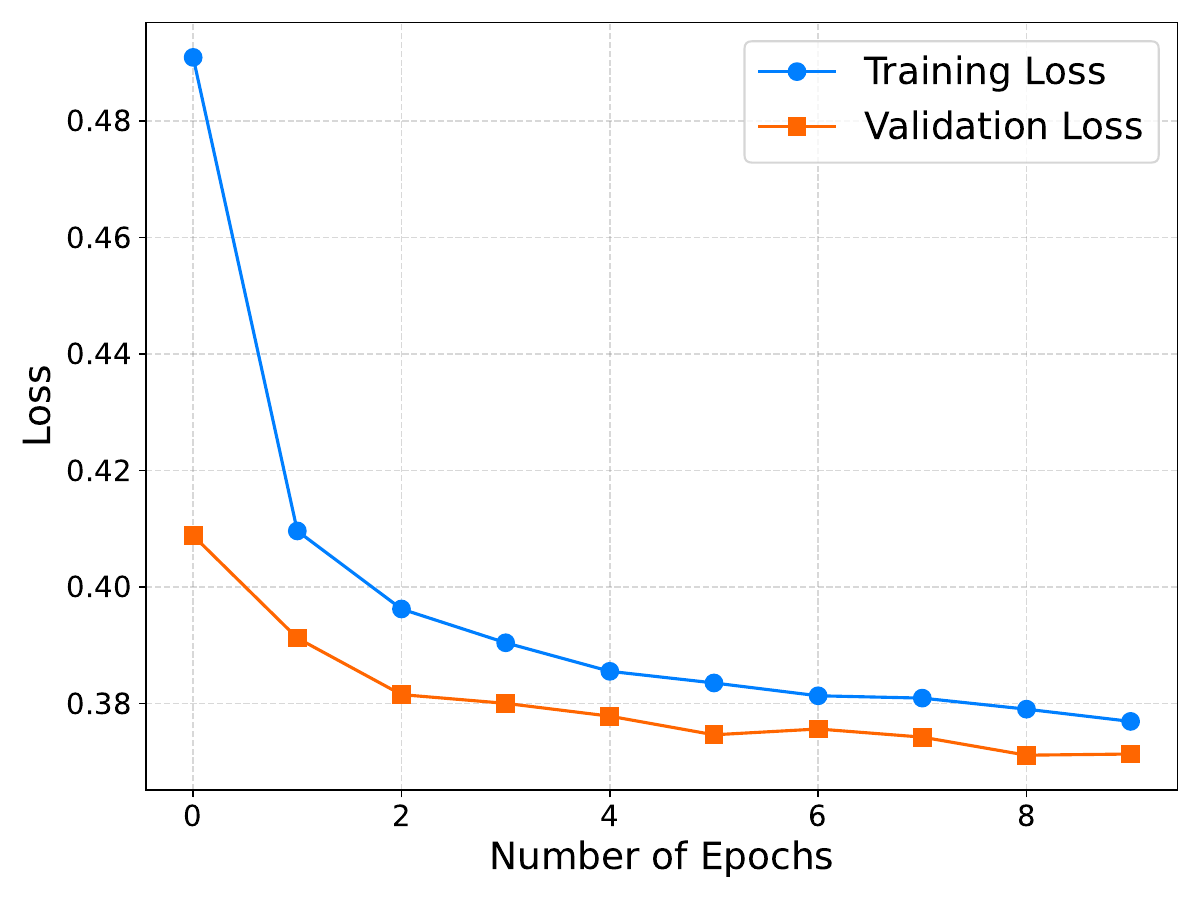}}\hfill

\caption{Graph representing model accuracy and model loss for training and validation set of proposed StrideNET model.}
\label{fig:acc}
\end{figure}

\subsection{Classification Results}

To evaluate the model, we use Overall Accuracy (OA), Average Accuracy (AA), and the Kappa Coefficient as primary metrics. Additionally, macro-averaged values of precision, recall, and F1 score are used for comprehensive evaluation.

Overall Accuracy represents the proportion of correctly classified instances across all classes relative to the total number of instances. Average Accuracy represents the mean accuracy across all classes.

The Kappa Coefficient is a robust statistical measure that evaluates the agreement between predicted and actual classifications, adjusted for chance agreement. Macro-averaged values of precision, recall, and F1 score are selected due to the balanced class distribution in our dataset.

The class-wise accuracy of the proposed StrideNET model is compared in Table \ref{tab:performance comparision} with standard CNN models, including MobileNet-V2, EfficientNet-B0, and ResNet-101, as well as transformer-based models such as ViT and DeiT. StrideNET achieves a test accuracy exceeding 99\% for each class.

Table \ref{tab:metrics} shows a comparison of StrideNET against other models using our primary metrics: Overall Accuracy (OA), Average Accuracy (AA), and the Kappa Coefficient (\(\kappa\)). Our proposed model also achieves values exceeding 99\% in these metrics.

To assess potential overfitting, we examined the accuracy-loss curves shown in Fig. \ref{fig:acc}. These curves, which illustrate the model's performance across varying epochs, indicate that the model maintains a balanced fit, avoiding both overfitting and underfitting. This empirical evidence suggests that our model is reliable and robust, and it can generalize well to unseen data.

In our comparative analysis (Table \ref{tab:metrics}), the StrideNET model demonstrated performance metrics on par with state-of-the-art models, including ResNet and DeiT. While both models achieved similar accuracy levels, StrideNET's efficiency in terms of data requirements and computational resources makes it the preferred choice for real-world terrain recognition tasks. 

\section{Conclusion}\label{conclusion}

This paper introduces StrideNET, a model based on Swin Transformer, to perform terrain recognition and surface roughness extraction from remote sensing images. It comprises of two branches: \textit{Terrain Recognition} and \textit{Roughness Extraction}. The model is trained to identify four terrain classes, namely grassy, marshy, rocky and sandy. A novel algorithm is proposed in this paper, which uses statistical texture-feature analysis to dynamically extract terrain characteristics of roughness and slipperiness from the input image. The model is trained on a custom dataset, and is benchmarked with standard convolutional and transformer based models. Experimental results demonstrate that the proposed model achieves classification accuracy of over $99\%$ for all terrain types, outperforming other models. Thus, it can be utilized for applications such as environmental monitoring, LULC classification, and precision agriculture.

\printbibliography
\end{document}